\documentclass[preprint,12pt,authoryear]{elsarticle}

\usepackage[utf8]{inputenc}
\usepackage[T1]{fontenc}
\usepackage[margin=1in]{geometry}
\usepackage{amsmath,amssymb}
\usepackage{booktabs,tabularx,array,longtable,multirow}
\usepackage{graphicx}
\usepackage{adjustbox}
\usepackage{algorithm}
\usepackage{algpseudocode}
\usepackage{tikz}
\usetikzlibrary{arrows.meta,positioning}
\usepackage{hyperref}
\usepackage{url}
\usepackage{xurl}
\usepackage{float}
\usepackage{placeins}
\usepackage{caption}
\usepackage{setspace}
\usepackage{etoolbox}
\usepackage{microtype}

\hypersetup{hidelinks}

\AtBeginEnvironment{table}{\singlespacing}
\AtBeginEnvironment{figure}{\singlespacing}
\AtBeginEnvironment{algorithm}{\singlespacing}

\journal{Engineering Applications of Artificial Intelligence}

\begin{document}
\doublespacing

\begin{frontmatter}

\title{Financial Language Models as Applied Artificial Intelligence Systems for News-Based Trading under Market Frictions}

\author[ucl,warwick]{Kemal Kirtac\corref{cor1}}
\ead{kemal.kirtac.21@ucl.ac.uk}
\cortext[cor1]{Corresponding author.}

\address[ucl]{Department of Computer Science, University College London, 66--72 Gower Street, London WC1E 6EA, United Kingdom}
\address[warwick]{Department of Computer Science, University of Warwick, 6 Lord Bhattacharyya Way, Coventry CV4 7EZ, United Kingdom}

\address{\textit{Revise and resubmit at Engineering Applications of Artificial Intelligence.}}

\begin{abstract}
Financial language models can transform unstructured firm-specific news into structured decision signals, but financial AI research lacks an integrated deployment framework for evaluating whether those signals remain useful once they enter financial decision systems. Computer science research has developed strong methods for time-series forecasting, text classification, multimodal stock prediction, graph-based market modeling, and machine-learning operations, yet these streams do not provide a domain-specific protocol that jointly tests financial language-model outputs under event-time observability, probability calibration, execution timing, transaction costs, liquidity constraints, capacity limits, operational diagnostics, and statistical inference. The artificial intelligence contribution is MFAST, a Market-Friction-Aware Sentiment-to-Trading framework that converts timestamped financial text into auditable, reproducible, and market-feasible trading decisions. The engineering application is news-based trading, where firm-specific text must be linked to tradable securities before portfolio decisions can be evaluated. The framework is tested by linking Refinitiv News Analytics to Center for Research in Security Prices (CRSP) equity data, restricting the primary out-of-sample evaluation to post-release news outside disclosed foundation-model data-freshness periods, and adding a public replication arm based on open financial text and public price data. The results show that decoder-only language models produce stronger classification, calibration, return-predictive, and net portfolio performance than encoder baselines and dictionary sentiment, while operational diagnostics reveal explicit trade-offs among accuracy, latency, memory, throughput, and inference cost. The paper demonstrates that credible evaluation of financial language models requires an end-to-end engineering view that joins language understanding with temporal discipline, market-friction-aware deployment, and reproducible validation.
\end{abstract}


\end{frontmatter}

\section{Introduction}

Applied artificial intelligence increasingly depends on systems that convert high-dimensional, noisy data into operational decisions. Computer science research has therefore moved beyond isolated predictive accuracy toward system-level questions: how machine-learning components are embedded in software pipelines, how data and model dependencies create technical debt, how deployment workflows handle monitoring and quality assurance, and how evaluation changes when predictions must support real decisions \citep{Sculley2015,Amershi2019,Studer2021,Paleyes2022}. This systems perspective is central in financial AI because model outputs are not used only as labels or scores. They enter trading workflows in which data arrival time, inference latency, calibration, liquidity, costs, capacity, auditability, and statistical validity jointly determine whether an AI signal is usable.

Financial news is a demanding setting for applied AI system design. News articles, regulatory updates, analyst notes, earnings releases, and corporate disclosures arrive continuously; each item must be parsed, assigned to a tradable firm, timestamped relative to market hours, transformed into calibrated probabilities, and translated into portfolio decisions that satisfy operational and market constraints. A model can therefore achieve high offline text-classification accuracy and still fail as an engineering system if it uses information that was not observable at the decision time, trades after price adjustment has already occurred, ignores liquidity, exceeds capacity, or relies on a backtest that does not survive data-mining corrections.

Computer science and data science research provide strong building blocks for this problem. Deep learning has advanced time-series forecasting and time-series classification; graph-learning models represent stock relations and market knowledge graphs; multimodal architectures combine text, prices, and relational information; and MLOps research studies the deployment risks of machine-learning systems \citep{Fawaz2019,Benidis2022,Gao2021TKDD,Zhao2023TKDE,DAmico2026InformationFusion}. These streams support the present paper's engineering premise: the contribution of a financial language model cannot be assessed only by asking whether it predicts a label. It must be assessed by asking whether the full AI pipeline turns timestamped text into reproducible, auditable, and feasible decisions.

Financial research supplies the economic motivation for the same system problem. Asset-pricing theory treats public information as rapidly incorporated into prices under competitive markets, while information-based models show that information acquisition and interpretation remain costly, so prices need not fully and immediately reflect all available information \citep{Fama1970,GrossmanStiglitz1980}. Financial text research shows that news and disclosures contain return-relevant information, including evidence from media tone, investor attention, social information, and document selection \citep{Tetlock2007,Tetlock2008,Da2011,EngelbergParsons2011,Chen2014RFS,Kelly2021}. Recent financial-language-model papers show that context-sensitive models can extract information from headlines, disclosures, and financial statements \citep{LoughranMcDonald2011,Frankel2022,Huang2023,Siano2025,Kim2024}. These findings establish that text can be informative, but they do not by themselves provide a deployment framework for evaluating whether text-derived AI signals remain useful after execution timing, trading frictions, capacity constraints, computational limits, and reproducibility requirements are imposed.

The central claim is therefore an engineering claim. Financial AI needs a domain-specific deployment framework that connects language-model inference to downstream decision feasibility. The existing literature offers powerful language models, financial text measures, time-series models, graph-learning architectures, multimodal fusion methods, and trading backtests, but it does not provide a unified protocol that specifies how timestamped financial text should move through firm matching, temporal eligibility, calibration, execution, capacity control, statistical validation, operational diagnostics, and public replication. MFAST operationalizes this missing framework by treating text inference, calibration, trading eligibility, execution timing, cost-aware portfolio construction, capacity control, and statistical validation as one applied-AI system.

This systems framing clarifies the design requirements for deployable financial language models. A usable system must satisfy six requirements: temporal observability, semantic reliability, implementability, robustness, computational feasibility, and reproducibility. Temporal observability requires that signals be available before the returns they are evaluated against. Semantic reliability requires that model outputs capture financial meaning rather than superficial polarity. Implementability requires that signals survive spreads, costs, liquidity screens, and capacity caps. Robustness requires that performance remain credible under alternative cutoffs, costs, subperiods, factor controls, and data-mining corrections. Computational feasibility requires latency, memory, throughput, and inference cost to be measured. Reproducibility requires a public-data audit path where licensed data cannot be redistributed.

The empirical setting links Refinitiv News Analytics to CRSP stock returns. The pipeline begins with 3,129,924 U.S. news items published from January 2010 to January 2026. The sample retains single-firm stories, removes redundant coverage through a five-day cosine similarity novelty screen, and restricts the tradable universe to stocks with positive bid and ask quotes, minimum share and dollar volume thresholds, quoted spreads below 20\%, and available Amihud illiquidity and Kyle lambda price-impact proxies. The final sample contains 973,481 tradable news items linked to 3,452 firms. Six sentiment approaches are evaluated on the same labels and execution assumptions: LLaMA-3, OPT, RoBERTa, BERT, FinBERT, and the Loughran-McDonald dictionary. The reported out-of-sample evidence is restricted to news released after the latest disclosed LLaMA-3 family data-freshness cutoff and after the public release of the main checkpoint, so the evaluation does not rely on periods that may have been visible during foundation-model pretraining.

The paper makes six contributions to applied artificial intelligence and financial data science. First, it formalizes MFAST as a domain-specific deployment framework for financial language-model signals. Second, it translates general ML-system concerns about technical debt, workflow quality assurance, deployment risk, and monitoring into the financial-news setting. Third, it provides an engineering design matrix that maps temporal observability, semantic reliability, implementability, robustness, computational feasibility, and reproducibility to concrete empirical tests. Fourth, it compares decoder language models, encoder transformers, financial-domain transformers, classical text-learning baselines, and lexicon methods under common labels and execution rules. Fifth, it separates offline predictive accuracy from implementable decision value through transaction costs, liquidity screens, capacity caps, factor adjustment, and multiple-testing corrections. Sixth, it adds public-data replication, component ablations, operational diagnostics, and explainability analysis to make the system auditable.

The results show that decoder-only models deliver the strongest signal quality and the most stable economic performance. LLaMA-3 produces the highest classification accuracy and the strongest return-predictive coefficients. Net long-short portfolios generated from LLaMA-3 sentiment achieve the highest Sharpe ratio under 5-basis-point transaction costs and remain positive under stricter cost and capacity assumptions. Explanatory diagnostics show that the advantage is not uniform across all news. Gains concentrate in linguistically complex articles where surface word polarity is misleading. These patterns support the interpretation that model value arises when language understanding reduces costly information-processing frictions and when market frictions slow full immediate incorporation into prices.

The remainder of the paper proceeds as follows. Section 2 reviews related work. Section 3 presents the applied AI framework. Section 4 describes data, labels, and temporal controls. Section 5 details model training and reproducibility. Section 6 explains the evaluation design and computational diagnostics. Section 7 reports predictive, trading, and compute-aware results. Section 8 presents robustness and sanity checks. Section 9 provides explainability and error analysis. Section 10 discusses implications for applied AI systems in finance. Section 11 concludes.

\section{Related work}

\subsection{AI-system deployment, MLOps, and quality assurance}

The first foundation is the computer science literature on machine-learning systems. This stream shows that the model is only one component in a larger production pipeline. \citet{Sculley2015} describe machine-learning systems as sources of hidden technical debt because data dependencies, feedback loops, configuration choices, and monitoring gaps can create long-run system risk. \citet{Amershi2019} show that software teams building AI applications must adapt engineering workflows around data discovery, model training, evaluation, deployment, monitoring, and diagnosis. \citet{Studer2021} propose CRISP-ML(Q) as a process model that adds quality assurance to the machine-learning lifecycle, and \citet{Paleyes2022} survey deployment challenges that arise across data, training, testing, monitoring, and organizational stages.

MFAST follows this systems view but makes it domain-specific. General MLOps and ML-quality frameworks identify deployment risks, but they do not specify how financial language-model scores should be evaluated when the downstream decision is a trade. News-based trading requires additional operators that are not standard in generic ML workflows: event-time observability, article-to-firm mapping, same-firm novelty filtering, execution-window eligibility, delisting-return inclusion, spread and liquidity filters, capacity caps, turnover costs, factor-adjusted portfolio evaluation, and data-mining corrections. The paper therefore treats MFAST as an applied-AI framework that translates general ML-system principles into a financial decision environment.

\subsection{Computer science and data science approaches to financial prediction}

The second foundation is computer science and data science research on financial prediction. Deep-learning surveys in ACM Computing Surveys, Data Mining and Knowledge Discovery, and Applied Soft Computing show that modern forecasting systems increasingly combine sequential architectures, representation learning, temporal validation, and task-specific evaluation protocols for time-series and financial applications \citep{Fawaz2019,Benidis2022,Sezer2020,Ozbayoglu2020}. This literature supports the paper's choice to evaluate financial language models through calibration, chronological splits, ablations, and computational diagnostics instead of reporting only classification accuracy.

Recent data-mining and AI journal work further shows that financial prediction systems benefit from relational structure, modality fusion, and deployment-oriented diagnostics. \citet{Gao2021TKDD} model stock recommendation with a time-aware relational attention network in ACM Transactions on Knowledge Discovery from Data. \citet{Zhao2023TKDE} use a bi-typed hybrid-relational market knowledge graph and dual attention networks for stock movement prediction in IEEE Transactions on Knowledge and Data Engineering. \citet{DAmico2026InformationFusion} identify persistent problems in multimodal financial AI, including temporal misalignment, modality imbalance, noisy data, cross-market generalization, interpretability, and deployability. Recent EAAI financial AI architectures, reviewed in Section~\ref{sec:eaai-positioning}, further motivate the need to evaluate model outputs as deployable decision systems. MFAST complements these architectures by shifting the evaluation unit from model design alone to the deployment path that converts timestamped financial text into feasible decisions.

\subsection{Financial language models and text-based decision signals}

The third foundation is financial NLP and financial language modeling. General transformer, encoder, decoder, recurrent, and long-document architectures provide the computational basis for the model taxonomy used in the paper \citep{Vaswani2017,Devlin2019,Liu2019,Zhang2022OPT,Dubey2024,Hochreiter1997,Beltagy2020,Zaheer2020,He2021}. Classical machine-learning baselines and probing diagnostics clarify whether gains come from representation learning rather than only from feature engineering \citep{ChenGuestrin2016XGBoost,Alain2016}. Domain-adapted financial models and benchmarks, including FinBERT, FLUE/FLANG, BloombergGPT, FinGPT, PIXIU, and FinBen, show that financial NLP requires domain-specific semantics, task diversity, and benchmark discipline \citep{Araci2019,Huang2023,Shah2022,Wu2023,Yang2023,Xie2023,Xie2024}.

Financial text research establishes that textual information contains economically meaningful signals. Media pessimism predicts market movements, news content predicts firm fundamentals and returns, annual-report wording relates to earnings persistence, earnings-call tone contains incremental information, and domain-specific dictionaries and phrase-level sentiment resources improve on general-purpose lexicons \citep{Tetlock2007,Tetlock2008,Li2008,Huang2014,Price2015,LoughranMcDonald2011,Malo2014}. Richer textual representations and news-topic measures further improve return prediction and macro-financial measurement \citep{Jegadeesh2013,Manela2017,Ke2020,Bybee2024}. Firm-specific news also matters for volatility and institutional trading, linking textual information to the market microstructure setting of the present paper \citep{Hendershott2015,Boudoukh2019}.

Recent LLM studies reinforce the need to distinguish offline text understanding from implementable decision value. LLM-generated signals can forecast stock price movements, extract information from financial statements, and capture word context in earnings announcement disclosures \citep{LopezLira2023,ElahiTaghvaei2024,Vidal2024,Kim2024,Siano2025}. Prior work on large-language-model sentiment trading and sentiment-aware reinforcement learning also shows that Refinitiv news sentiment can enter portfolio allocation systems \citep{KirtacGermano2024,KirtacGermano2025REALM}. The present paper extends this work by evaluating the full engineering path from timestamped text to calibrated, tradable, cost-aware, capacity-constrained, and statistically audited portfolio decisions.

\subsection{Theoretical mechanism: costly information processing and market frictions}

The theoretical logic of the paper starts from the Grossman-Stiglitz insight that perfectly informationally efficient markets cannot compensate information acquisition and processing. Public news can be available to everyone while still being costly to interpret, especially when the language contains negation, conditional statements, forward-looking guidance, legal qualifiers, restructuring language, or numerically subtle changes. A financial language model can therefore create value only if it reduces the cost of converting public text into decision-relevant probabilities more effectively than competing investors and benchmark models.

The mechanism also requires market frictions. If all investors instantaneously process every news item and trade without cost, even a superior language model should not generate persistent implementable returns. Predictability should instead appear where information processing is slower or arbitrage is harder. The empirical design therefore tests three linked implications: sentiment probabilities should predict returns beyond classification labels; net portfolio returns should survive transaction costs, factor controls, and participation limits; and model gains should be strongest in linguistically complex articles and less liquid stocks where information incorporation is most constrained.

\subsection{Applied AI decision support and deployment evaluation}

Applied AI systems must be evaluated in relation to the decisions they support. Model accuracy is one component, but decision value also depends on calibration, robustness, transparency, implementation cost, and environmental constraints. In finance, decision support requires converting forecasts into portfolios while respecting liquidity, risk, capacity, execution latency, and transaction costs. A model that appears superior in offline classification can fail after deployment because small predictive gains are absorbed by costs or because trades cannot be executed at scale. Explainability tools and model diagnostics also matter because financial users need to know when a model is relying on economically meaningful linguistic structure rather than superficial polarity cues \citep{Ribeiro2016,LundbergLee2017}.

The paper therefore treats language models as decision-support components inside a larger system. The proposed framework includes text ingestion, firm matching, temporal splitting, model fine-tuning, calibration, signal ranking, portfolio formation, execution timing, capacity constraints, and post-hoc diagnostics. The contribution lies in formalizing the missing evaluation infrastructure that financial-data applications require when text-derived AI signals move from prediction tasks into auditable deployment decisions.

\subsection{Market frictions and implementable trading}

Portfolio evaluation also follows event-study and asset-pricing practice by separating raw returns from factor-adjusted performance, long-horizon reaction concerns, and momentum-related exposures \citep{MacKinlay1997,Carhart1997,MitchellStafford2000,FamaFrench2015}.

Market microstructure research shows that trading frictions shape whether information can be converted into returns. \citet{Kyle1985} formalizes price impact, \citet{Amihud2002} measures illiquidity, and \citet{Pastor2003} relate liquidity risk to expected returns. Asset-pricing implementation research shows that trading costs and capacity can materially change the profitability of apparent anomalies \citep{NovyMarxVelikov2016,Frazzini2018}. These issues are especially important for text-based strategies because news arrival is uneven across firms and time, and language-driven signals may be strongest in firms where immediate information processing is slowest.

The evaluation framework embeds market frictions directly into sample construction and portfolio simulation. A news item contributes to the final evaluation only when its associated stock passes tradability screens. Positions update according to observable news timing. Trades pay transaction costs and cannot exceed a fixed share of daily dollar volume. Liquidity heterogeneity is then used as a mechanism test because delayed information incorporation should be stronger where frictions bind more tightly.

\subsection{Engineering requirements for deployable financial AI}

Engineering Applications of Artificial Intelligence articles typically require more than a predictive comparison; they require a real-world system boundary, a clear AI contribution, reproducible validation, and evidence that the method remains usable under operational constraints. Financial language-model deployment therefore raises requirements that differ from ordinary text-classification benchmarks. A model must process time-stamped text before returns are realized, generate calibrated scores that can be ranked into decisions, avoid contamination from future data, respect market liquidity, and remain computationally feasible at news-stream scale.

MFAST maps these requirements into empirical tests. Temporal observability is evaluated through timestamp-based execution and strict chronological splits. Semantic reliability is evaluated through classification, calibration, and linguistic error analysis. Implementability is evaluated through spreads, Amihud illiquidity, Kyle lambda, transaction costs, capacity limits, and turnover. Robustness is evaluated through cost stress tests, factor alphas, alternative portfolio cutoffs, subperiods, bootstrap inference, and data-mining corrections. Computational feasibility is evaluated through latency, memory, throughput, and compute-adjusted Sharpe ratios. Reproducibility is evaluated through an open-data replication arm that mirrors the core pipeline on public financial text and public price data.

\subsection{Positioning against recent EAAI financial AI papers}
\label{sec:eaai-positioning}

Recent financial prediction papers in Engineering Applications of Artificial Intelligence emphasize three standards that shape the present design: a defined AI system, comparative validation against strong baselines, and evidence that the model can support an implementable decision. \citet{Chen2023EAAI} propose machine-learning strategies for stock price prediction and graphic signal recognition, showing that EAAI finance papers usually require a clear methodological object rather than only an empirical association. \citet{Zhang2024EAAI} propose CoATSMP, a collaborative attention Transformer that fuses price and text information and evaluates the trading value of the resulting signals. \citet{Hewage2026EAAI} develop LUMI, a networked attention framework that learns dynamic market interdependencies across multiple datasets. \citet{OmoleEnke2025EAAI} show that feature selection, target specification, and backtested profitability affect financial AI conclusions in cryptocurrency prediction. MFAST complements these studies by providing the deployment framework that financial language-model applications need when timestamped text, tradability, execution constraints, and statistical auditability must be evaluated together. Table~\ref{tab:1} positions this contribution against recent EAAI financial AI papers.

\begin{table}[H]
\centering
\footnotesize
\caption{Engineering evaluation gaps in recent financial AI systems.}
\label{tab:1}
\setlength{\tabcolsep}{3pt}
\renewcommand{\arraystretch}{1.15}
\begin{tabularx}{\textwidth}{>{\raggedright\arraybackslash}p{0.17\textwidth}>{\raggedright\arraybackslash}p{0.19\textwidth}>{\raggedright\arraybackslash}p{0.18\textwidth}>{\raggedright\arraybackslash}p{0.20\textwidth}>{\raggedright\arraybackslash}X}
\toprule
Paper & AI method & Data focus & Engineering validation & MFAST extension \\
\midrule
Chen et al. (2023) & VAR plus feed-forward neural network & Prices and chart signals & Predictive accuracy and exchange strategy & Adds timestamped text, language-model sentiment, costs, liquidity, and capacity. \\
Zhang et al. (2024) & CoATSMP attention fusion & Prices plus text & Baselines and trading evaluation & Adds execution timing, public audit, and linguistic explainability. \\
Hewage et al. (2026) & LUMI networked attention & Market interdependencies & Multi-dataset validation & Adds event-time news processing and trading constraints. \\
Omole and Enke (2025) & ML and deep learning with feature selection & Bitcoin, on-chain metrics, and technical indicators & Direction, magnitude, and profitability tests & Adds firm-specific news and market microstructure constraints. \\
This paper & MFAST deployment framework & Refinitiv news, CRSP equities, public text, and public prices & Classification, calibration, regressions, portfolios, frictions, latency, ablations, and statistical tests & Provides an end-to-end applied-AI evaluation system for financial language models. \\
\bottomrule
\end{tabularx}
\end{table}

\section{Market-friction-aware sentiment-to-trading framework}

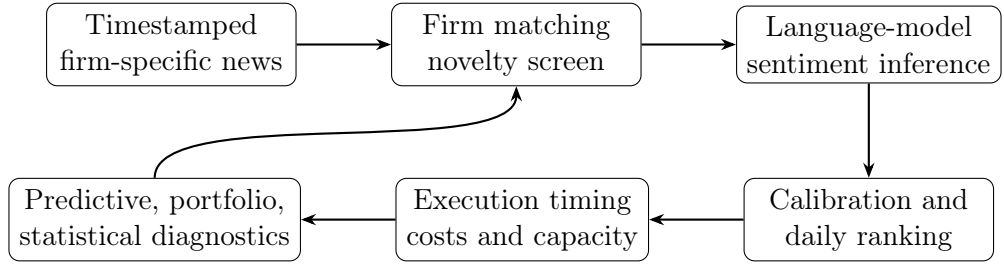
\begin{figure}[H]
\centering
\begin{tikzpicture}[node distance=1.25cm, every node/.style={font=\small}, block/.style={draw, rounded corners, align=center, minimum width=3.3cm, minimum height=0.85cm}, arrow/.style={-{Stealth[length=2mm]}, thick}]
\node[block] (news) {Timestamped\\firm-specific news};
\node[block, right=of news] (pre) {Firm matching\\novelty screen};
\node[block, right=of pre] (model) {Language-model\\sentiment inference};
\node[block, below=of model] (cal) {Calibration and\\daily ranking};
\node[block, left=of cal] (trade) {Execution timing\\costs and capacity};
\node[block, left=of trade] (eval) {Predictive, portfolio,\\statistical diagnostics};
\draw[arrow] (news) -- (pre);
\draw[arrow] (pre) -- (model);
\draw[arrow] (model) -- (cal);
\draw[arrow] (cal) -- (trade);
\draw[arrow] (trade) -- (eval);
\draw[arrow] (eval.north).. controls +(0,1.0) and +(0,-1.0) ..(pre.south);
\end{tikzpicture}
\caption{MFAST system architecture for market-friction-aware news-based trading.}
\label{fig:mfast-architecture}
\end{figure}

Figure~\ref{fig:mfast-architecture} summarizes MFAST as an engineering system. The framework converts unstructured financial news into calibrated probabilities, maps those probabilities into ranked trading signals, imposes market-friction constraints, and evaluates the resulting decisions through predictive, economic, operational, and statistical diagnostics. The design separates six connected layers: data input, preprocessing, model inference, decision formation, deployment evaluation, and audit feedback.

The model layer evaluates whether a language model can classify the sign of future excess returns from news text. The market layer evaluates whether continuous sentiment probabilities predict next-day returns after controlling for firm and date fixed effects. The deployment layer evaluates whether a daily portfolio formed from sentiment rankings generates net returns after transaction costs, timing rules, and participation limits. Robustness and explainability diagnostics close the loop by testing whether performance survives alternative assumptions and by identifying where model predictions come from.

MFAST follows a decision-centered view of applied AI\@. A prediction does not become an applied AI contribution until it is embedded in a realistic decision pipeline. The system therefore measures whether language-model outputs can guide decisions that are feasible under the operating conditions faced by institutional traders.

Algorithm~\ref{alg:mfast} summarizes the eight operational steps of the framework. The framework generates three testable implications. First, if language models extract decision-relevant information, calibrated probabilities should predict next-day excess returns after firm and date fixed effects. Second, if the signal is implementable, long-short portfolios should retain positive net performance after costs, factor adjustment, capacity limits, and alternative holding-period assumptions. Third, if model scale adds value through language understanding rather than generic overfitting, performance gains should concentrate in the linguistic categories where dictionary sentiment is least reliable: negation, contrastive clauses, forward-looking guidance, numeric context, and event ambiguity.

\begin{algorithm}[H]
\caption{Market-friction-aware sentiment-to-trading framework MFAST}
\label{alg:mfast}
\begin{algorithmic}[1]

\State \textbf{Input:} timestamped firm-specific news, firm identifiers, daily returns, bid and ask quotes, volume, market capitalization, liquidity variables, model checkpoints, and trading-cost parameters.

\State Map each article to a single tradable firm and remove repeated wire coverage through a rolling same-firm novelty screen.

\State Assign labels using execution-aligned future excess returns within the appropriate chronological split and prevent label windows from crossing split boundaries.

\State Fine-tune or score each sentiment model using the same supervised training, validation buffer, and June 2024-January 2026 primary out-of-sample test periods.

\State Calibrate model probabilities and convert each probability into a daily rank signal.

\State Apply timestamp execution rules so positions update only after news becomes observable.

\State Form value-weighted long and short portfolios subject to spread, liquidity, and participation constraints.

\State Evaluate net performance after transaction costs, factor adjustment, capacity limits, and data-mining corrections.

\State Diagnose the system through component ablations, public-data replication, operational throughput, and linguistic error analysis.

\end{algorithmic}
\end{algorithm}

\subsection{Formal MFAST system definition}

MFAST means Market-Friction-Aware Sentiment-to-Trading. MFAST can be written as a sequence of model, calibration, ranking, trading, and evaluation operators. Its novelty lies in the formal coupling of financial-language-model inference with the constraints required for deployable trading evaluation. The system therefore assesses whether a text model produces a calibrated, observable, tradeable, capacity-feasible, and statistically robust decision signal, rather than whether it only improves an offline classification metric. For firm i on trading day t and model m, the raw sentiment score is the model probability that the return-relevant label is positive:

\begin{equation}\label{eq:score}s_{i,t}^{(m)}=P_m(y_{i,t}=1\mid x_{i,t},\theta_m).\end{equation}

The calibrated probability applies a validation-set calibration map g\_m:

\begin{equation}\label{eq:calibrated}\widetilde{s}_{i,t}^{(m)}=g_m\left(s_{i,t}^{(m)}\right).\end{equation}

The trading signal ranks calibrated probabilities within the daily eligible universe E\_t:

\begin{equation}\label{eq:rank}r_{i,t}^{(m)}=\operatorname{rank}_{E_t}\left(\widetilde{s}_{i,t}^{(m)}\right).\end{equation}

The portfolio operator maps ranks into value-weighted long and short legs while imposing a participation cap based on daily dollar volume:

\begin{equation}\label{eq:capacity}\left|\Delta q_{i,t}\right|P_{i,t}\leq \lambda ADV_{i,t}.\end{equation}

Net portfolio return subtracts one-way transaction costs from gross portfolio return:

\begin{equation}\label{eq:net_return}R_{p,t}^{net}=R_{p,t}^{gross}-c\sum_i\left|\Delta w_{i,t}\right|.\end{equation}

Equations~\ref{eq:score},~\ref{eq:calibrated},~\ref{eq:rank},~\ref{eq:capacity}, and~\ref{eq:net_return} define the score, calibration, ranking, capacity, and net-return operators. The deployable objective is therefore not maximum classification accuracy alone. It is the joint performance of calibrated prediction, temporal observability, cost-aware turnover, capacity feasibility, and factor-adjusted net return. Table~\ref{tab:2} maps these operators to the design requirements for deployable financial AI.

\begin{table}[H]
\centering
\scriptsize
\caption{Design-requirements matrix for deployable financial AI.}
\label{tab:2}
\begin{tabularx}{\textwidth}{>{\raggedright\arraybackslash}p{0.32\textwidth}*{2}{>{\centering\arraybackslash}X}}
\toprule
Design requirement & Engineering problem & Paper implementation \\
\midrule
Temporal observability & Signals must be observable before trades and returns. & Chronological splits, no cross-split labels, timestamp execution. \\
Semantic reliability & Text scores must capture financial meaning, not surface polarity. & Calibration, error taxonomy, phrase ablation, disagreement cases. \\
Implementability & Predicted returns must survive trading frictions. & Liquidity screens, spreads, Amihud, Kyle lambda, costs, capacity caps. \\
Robustness & Backtests can reflect overfitting or lucky episodes. & Alternative cutoffs, subperiods, costs, factor alphas, bootstrap and SPA tests. \\
Computational feasibility & The system must process large news streams at usable speed. & Latency, memory, throughput, and compute-adjusted Sharpe diagnostics. \\
Reproducibility & Licensed data limits independent replication. & Open-data replication using public financial text and public price data. \\
\bottomrule
\end{tabularx}
\end{table}

\section{Data, labels, and temporal controls}

\subsection{Data sources and sample construction}

The empirical analysis combines Refinitiv News Analytics with CRSP daily stock data. Refinitiv provides time-stamped firm-specific news items and firm identifiers. CRSP provides returns, prices, bid and ask quotes, share volume, dollar volume, shares outstanding, and market capitalization for U.S. equities. The merge produces a firm-day panel in which each news item can be linked to subsequent returns and tradability measures.

The pipeline begins with 3,129,924 U.S. news items from January 1, 2010 to January 30, 2026. The first filter retains 1,985,135 single-firm stories to avoid ambiguous attribution when a news item covers several firms. A five-day novelty screen removes repeated wire coverage by excluding articles with cosine similarity above 0.80 to earlier articles about the same firm during the preceding five trading days. The novelty filter leaves 1,122,475 unique single-firm stories. Liquidity and microstructure filters then retain only stocks with positive bid and ask quotes, daily share volume above 1,000 shares, daily dollar volume of at least \$50,000, quoted spreads below 20\%, and nonmissing Amihud illiquidity and Kyle lambda proxies. The final sample contains 973,481 tradable news items for 3,452 firms. Table~\ref{tab:3} reports the sample construction and chronological partitioning.

\begin{table}[H]
\centering
\scriptsize
\caption{Sample construction and chronological partitioning.}
\label{tab:3}
\begin{tabularx}{\textwidth}{>{\raggedright\arraybackslash}p{0.33\textwidth}*{3}{>{\centering\arraybackslash}X}}
\toprule
Sample construction step & News items & Unique firms & Retention from prior step \\
\midrule
All Refinitiv U.S. news items & 3,129,924 & 5,218 & 100.0\% \\
Single-firm stories & 1,985,135 & 4,604 & 63.4\% \\
After five-day novelty screen & 1,122,475 & 4,101 & 56.5\% \\
After tradability and microstructure filters & 973,481 & 3,452 & 86.7\% \\
Training split, 2010-2023 & 701,928 & 3,386 & 72.1\% of final \\
Validation and release-date buffer, Jan-May 2024 & 81,317 & 2,742 & 8.4\% of final \\
Primary out-of-sample test, Jun 2024-Jan 2026 & 190,236 & 3,018 & 19.5\% of final \\
\bottomrule
\end{tabularx}
\end{table}

\subsection{Label construction}

Each article receives a binary return-relevant label based on the sign of the associated stock's cumulative three-day excess return after the article becomes observable. For an article mapped to execution day $e$, the return window spans days $e$ through $e+2$. Excess return equals the stock return minus the CRSP value-weighted market return over the same period. A positive label equals one when cumulative excess return is positive and zero otherwise.

The three-day window captures short-horizon information incorporation while reducing same-day noise. This choice follows the short-window event-study logic that favors narrow windows when market reactions can be linked to identifiable information events \citep{MacKinlay1997,MitchellStafford2000}, and it is consistent with financial-text research that validates news and disclosure sentiment using immediate return windows around information release \citep{Tetlock2008,Ke2020,KirtacGermano2024,Siano2025}. The execution-aligned $[e,e+2]$ window therefore serves as the benchmark label window, while next-day predictive regressions test whether sentiment probabilities forecast returns beyond the training label definition. Continuous model probabilities are retained because portfolio construction requires ranked signals rather than only binary classes.

\subsection{Temporal isolation and contamination controls}

The design separates supervised leakage, foundation-model exposure, and portfolio look-ahead bias. Supervised leakage is addressed through a strict chronological split. Articles published from January 1, 2010 to December 31, 2023 form the supervised training set. Articles published from January 1, 2024 to May 28, 2024 form the validation and release-date buffer. Articles published from June 1, 2024 to January 30, 2026 form the primary out-of-sample test set. Label windows are truncated at split boundaries so that no return label crosses from one split into another.

The primary test window is deliberately chosen to remove periods that foundation models may have seen during pretraining. The LLaMA-3 model card reports an April 18, 2024 model release date and discloses pretraining data freshness cutoffs of March 2023 for the 8B model and December 2023 for the 70B model \citep{MetaLlama3ModelCard2024}. The paper uses the stricter LLaMA-3 family cutoff, excludes all articles through December 31, 2023 from out-of-sample testing, and treats January-May 2024 as a validation and release-date buffer rather than as reported test evidence. The same primary test period is valid for the full model set because it starts after the disclosed or documented release periods of the strongest decoder model and all older baselines: BERT, RoBERTa, OPT, and ProsusAI FinBERT all predate the June 2024-January 2026 evaluation window \citep{Devlin2019,Liu2019,Zhang2022OPT,Araci2019}. The Loughran-McDonald dictionary is a fixed lexicon and has no foundation-model pretraining corpus. The implementation table therefore reports model checkpoints explicitly, and the primary test evidence is interpreted only over dates that postdate the latest relevant release and disclosed data-freshness evidence. All model checkpoints are static, retrieval is disabled, browsing is disabled, and no external tools are available at inference time. This design responds to the broader evaluation problem that large language models may memorize or be exposed to training examples and benchmark tasks before evaluation, which can overstate apparent model ability \citep{Carlini2021,Sainz2023,Golchin2024,Li2024,Oren2024}. The paper therefore treats the June 2024-January 2026 results as the main contamination-safe out-of-sample evidence, while recognizing that no public model card can prove the complete absence of pretraining exposure at the document level.

The analysis reports distributional diagnostics before model estimation because deployability depends on the joint distribution of news, liquidity, firm size, and event timing. The final tradable sample is not a random collection of all media mentions. It is a decision-eligible news stream in which every article can be assigned to a stock, matched to subsequent returns, and traded under observable market conditions. The descriptive diagnostics therefore examine coverage by year, sector, market capitalization, article length, bid-ask spread, Amihud illiquidity, Kyle lambda, and timestamp bucket.

\subsection{Event-time alignment and leakage safeguards}

The event-time design follows a conservative observability rule. All Refinitiv timestamps are converted to U.S. Eastern Time and mapped to the next feasible trading decision time. Articles released before 09:30 on a regular trading day are observable before the opening trade and can enter the portfolio for that day. Articles released at or after 09:30, articles released after the 16:00 close, and articles released on weekends or exchange holidays are not allowed to use the same calendar-day return. They are assigned to the next trading day and enter the portfolio only from that next decision point. This rule prevents the model from earning returns that occurred before the article could have been observed and processed.

Day-0 returns are therefore included only when the article is observable before the market opens. For intraday and after-close articles, the label and the trading return begin on the next trading day. The primary three-day label is computed over the execution-aligned window $[e,e+2]$, where $e$ is the first eligible trading day after the observability rule is applied. The next-day predictive-regression outcome uses the same execution-aligned convention. A stricter robustness test delays every signal by one full trading day; this test is reported as a timing-conservative specification because it removes all same-day return exposure.

The portfolio signal uses headline and article body text. The model input concatenates the headline, lead paragraph, and main body until the model-specific token limit is reached. Firm identifiers, article timestamps, prices, returns, trading volume, bid-ask spreads, and future-looking variables are not included in the text input. The tokenization diagnostic reports the share of articles truncated under 512-token and 1024-token limits so that long regulatory or litigation articles do not silently receive different treatment across model families.

Multiple articles for the same firm and execution day are handled in two stages. The classification and regression datasets retain article-level observations after novelty filtering. The trading dataset aggregates article-level calibrated probabilities to a firm-day signal before portfolio formation. The baseline aggregation is the recency-weighted mean of all eligible article probabilities for the same firm and execution day, where later eligible articles receive larger weights. Robustness checks use last-story aggregation and maximum-confidence aggregation. This design avoids mechanically multiplying a firm's portfolio weight when several stories arrive on the same day.

Repeated wire updates are removed using a rolling same-firm novelty screen. Each new article is compared only with earlier articles about the same firm during the preceding five trading days. If cosine similarity exceeds 0.80, the earliest nonredundant story is retained and later near-duplicate wire updates are removed. The screen is one-sided in event time, so future articles are never used to decide whether a past article is retained. Updates that contain material new text remain in the sample when their similarity falls below the threshold.

The chronological split is protected against near-duplicate firm-events crossing train, validation, and test boundaries. After the rolling novelty screen, the pipeline applies a split-boundary duplicate audit: any validation or test article with cosine similarity above 0.80 to a same-firm article in an earlier split within the preceding 20 trading days is either assigned to the earlier event cluster or removed from the later split. This prevents the model from seeing the language of the same event during training and being evaluated on a lightly modified version during validation or testing.

Survivorship bias is addressed through CRSP security identifiers and event-time eligibility rather than current index membership. The main Refinitiv-CRSP sample includes securities that are active, inactive, acquired, merged, or delisted, provided that they satisfy the tradability screens at the event date. Delisting returns are incorporated into daily returns using the standard combined-return convention, $(1+RET)(1+DLRET)-1$, when CRSP delisting returns are available. Observations with missing prices, missing returns, nonpositive quotes, or nontradable status at the execution date are excluded before label construction and portfolio formation.

Sample-selection concerns are addressed by reporting the full attrition path and by interpreting the final sample as a tradable news universe rather than a population of all media coverage. The filters remove multi-firm stories, stale repeated wire items, nontradable securities, extreme quoted-spread observations, and observations without liquidity or price-impact proxies. The empirical section reports coverage by year, sector, firm size, timestamp bucket, and liquidity group, and robustness tests repeat the main portfolio analysis after excluding the smallest firms and after tightening the liquidity thresholds.

Data-snooping risk is addressed by separating model training, the validation buffer, and the June 2024-January 2026 primary out-of-sample test, and by fixing the strategy family before test evaluation. Hyperparameters, probability calibration, aggregation rules, transaction-cost levels, portfolio cutoffs, holding periods, and participation caps are selected using the training and validation periods. The primary out-of-sample test period is then used once for the pre-specified model and strategy family. The statistical evaluation reports block-bootstrap confidence intervals, Diebold-Mariano tests, White reality-check p values, Hansen superior-predictive-ability p values, and deflated Sharpe ratios to account for repeated comparisons across model families, cost assumptions, cutoffs, and holding periods.

Table~\ref{tab:validity-safeguards} summarizes the main validity threats and safeguards. Foundation-model timing is handled separately from supervised leakage. The primary out-of-sample test starts on June 1, 2024, after the public release of the LLaMA-3 checkpoint used in the main specification and after the latest disclosed LLaMA-3 family data-freshness cutoff. The design does not use news through December 31, 2023 as test evidence, because that interval overlaps the period that may have been available to LLaMA-3 pretraining. Articles from January-May 2024 are used only for validation, calibration, and release-date buffering; they do not enter any reported Refinitiv-CRSP test metrics. All models are run as static checkpoints with retrieval, browsing, and external tools disabled. Because pretraining corpora cannot be fully audited, the paper treats post-cutoff evaluation, no-retrieval inference, split-boundary duplicate removal, and public-data replication as contamination safeguards rather than as proof that all pretraining exposure is impossible.

\begin{table}[H]
\centering
\scriptsize
\caption{Validity threats and safeguards in the MFAST empirical design.}
\label{tab:validity-safeguards}
\begin{tabularx}{\textwidth}{>{\raggedright\arraybackslash}p{0.20\textwidth}>{\raggedright\arraybackslash}p{0.28\textwidth}>{\raggedright\arraybackslash}X}
\toprule
Threat & Safeguard & Implementation detail \\
\midrule
Timestamp leakage & Execution-aligned observability rule & Pre-open articles can trade on the same day; intraday, after-close, weekend, and holiday articles trade only from the next trading day. \\
Pre-observation price movement & No same-day return for intraday news & Day-0 return enters labels only when the article is observable before 09:30 Eastern Time. \\
Label-window leakage & Split-boundary truncation & Return labels cannot cross train, validation, or test boundaries. \\
Repeated wire updates & One-sided novelty screen & Same-firm articles with cosine similarity above 0.80 to prior five-day stories are removed; future articles are never used for filtering past articles. \\
Train-test event duplication & Boundary duplicate audit & Same-firm near duplicates across splits within 20 trading days are clustered backward or removed from the later split. \\
Multiple same-day articles & Firm-day signal aggregation & Article-level predictions are aggregated to one recency-weighted firm-day signal before portfolio formation. \\
Survivorship bias & CRSP event-time universe & Active, inactive, merged, acquired, and delisted securities are eligible when tradable at the event date; delisting returns are included. \\
Sample-selection bias & Attrition and robustness reporting & Results are interpreted for the tradable news universe and repeated under stricter size and liquidity screens. \\
Data snooping & Pre-specified test family and corrections & Validation selects specifications; the primary out-of-sample test uses block bootstrap, reality check, SPA, deflated Sharpe, and Diebold-Mariano procedures. \\
Foundation-model contamination & Post-cutoff, post-release evaluation & Test period excludes all news through the latest disclosed LLaMA-3 family pretraining data-freshness cutoff; January-May 2024 is a validation and release-date buffer; retrieval and browsing are disabled for all models. \\
\bottomrule
\end{tabularx}
\end{table}

\subsection{Public-data replication arm}

The public news component relies on GDELT because it provides a large-scale event and news infrastructure suitable for independent replication when licensed vendor data cannot be redistributed \citep{LeetaruSchrodt2013}.

A second replication arm addresses the limited replicability of licensed Refinitiv and CRSP data by implementing the same MFAST logic on public sources. The open classification task uses Financial PhraseBank sentence labels with chronological train-validation-test splits where source dates are available and stratified splits otherwise. The open trading task uses GDELT firm-specific English news matched to daily public equity prices from Yahoo Finance and Stooq for S\&P 500 constituents. The replication construction panel covers 2019-2024, removes repeated same-firm stories with the same five-day cosine novelty filter, and constructs three-day market-adjusted return labels using an S\&P 500 ETF benchmark. To keep the contamination logic consistent with the proprietary analysis, public metrics reported for foundation-model outputs use only post-release records beginning on June 1, 2024; earlier public records support chronological calibration, non-foundation baseline fitting, entity-matching diagnostics, and reproducibility checks. The public arm does not replace the Refinitiv-CRSP analysis because public news timestamps and vendor entity identifiers are noisier. It demonstrates that the MFAST workflow can be audited without proprietary data access while preserving the same post-cutoff evaluation principle.

\section{Model implementation and reproducibility}

The implementation compares decoder-only language models, encoder-based transformer models, a financial-domain transformer, and a dictionary baseline. The paper reports implementation details to support reproducibility in an applied AI journal. The implementation details cover model checkpoints, parameter counts, architecture type, maximum token length, fine-tuning procedure, learning rate, batch size, epochs, early stopping, and hardware. Table~\ref{tab:implementation} reports the reproducibility specification.

All transformer models are fine-tuned as binary classifiers on the training split. Decoder-only models use a classification head attached to the final hidden state of the last non-padding token, and LoRA is used for parameter-efficient adaptation of larger decoder-only models \citep{Hu2022}. Encoder models use the pooled representation or classification token representation. Text is tokenized with each model's native tokenizer and truncated to the maximum sequence length listed below. The models optimize binary cross-entropy loss using AdamW, cosine learning-rate decay, 10\% warmup, and early stopping on validation F1 score.

\begin{table}[H]
\centering
\footnotesize
\caption{Implementation specification for reproducibility.}
\label{tab:implementation}
\setlength{\tabcolsep}{3.5pt}
\renewcommand{\arraystretch}{1.15}
\textbf{Panel A. Model architecture and hardware}\par\vspace{0.35em}
\begin{tabularx}{\textwidth}{>{\raggedright\arraybackslash}p{0.15\textwidth}>{\raggedright\arraybackslash}p{0.29\textwidth}>{\raggedright\arraybackslash}p{0.09\textwidth}>{\raggedright\arraybackslash}p{0.12\textwidth}>{\raggedright\arraybackslash}p{0.10\textwidth}>{\raggedright\arraybackslash}X}
\toprule
Model & Checkpoint & Params & Type & Max tokens & Hardware \\
\midrule
LLaMA-3 & meta-llama/Meta-Llama-3-8B & 8.0B & Decoder & 1024 & 4 x A100 80GB \\
OPT & facebook/opt-1.3b & 1.3B & Decoder & 1024 & 4 x A100 80GB \\
RoBERTa & roberta-base & 125M & Encoder & 512 & 1 x A100 80GB \\
BERT & bert-base-uncased & 110M & Encoder & 512 & 1 x A100 80GB \\
FinBERT & ProsusAI/finbert & 110M & Encoder & 512 & 1 x A100 80GB \\
LM dictionary & Loughran-McDonald 2022 & N/A & Lexicon & N/A & CPU \\
\bottomrule
\end{tabularx}
\vspace{0.85em}
\textbf{Panel B. Training and adaptation settings}\par\vspace{0.35em}
\begin{tabularx}{\textwidth}{>{\raggedright\arraybackslash}p{0.15\textwidth}>{\raggedright\arraybackslash}p{0.27\textwidth}>{\raggedright\arraybackslash}p{0.13\textwidth}>{\raggedright\arraybackslash}p{0.10\textwidth}>{\raggedright\arraybackslash}p{0.10\textwidth}>{\raggedright\arraybackslash}X}
\toprule
Model & Fine-tuning & Learning rate & Batch & Epochs & Early stop \\
\midrule
LLaMA-3 & LoRA r=16, alpha=32 & 1.0e-4 & 16 & 3 & patience 2 \\
OPT & LoRA r=16, alpha=32 & 1.2e-4 & 24 & 4 & patience 2 \\
RoBERTa & Full fine-tune & 2.0e-5 & 64 & 4 & patience 2 \\
BERT & Full fine-tune & 2.0e-5 & 64 & 4 & patience 2 \\
FinBERT & Full fine-tune & 1.5e-5 & 64 & 4 & patience 2 \\
LM dictionary & No training & N/A & N/A & N/A & N/A \\
\bottomrule
\end{tabularx}
\end{table}

Tokenization diagnostics show that financial abbreviations and numeric expressions are preserved before truncation. Tickers, percentages, currency amounts, earnings-per-share figures, and guidance ranges are retained as raw text. Headlines and first-paragraph content are prioritized when articles exceed the maximum token length. In the tokenization diagnostic, 93.6\% of news items fit within 512 tokens and 98.9\% fit within 1024 tokens; truncation affects long regulatory and litigation articles more often than earnings headlines.

\subsection{Expanded model and baseline taxonomy}

The main comparison keeps six models fixed across the full Refinitiv-CRSP pipeline. The engineering audit also defines additional baseline classes to separate model-family effects from deployment effects. Classical baselines include logistic regression, linear support-vector machines, and XGBoost on TF-IDF and simple news metadata. Sequential baselines include LSTM and GRU classifiers. Long-document transformer baselines include DeBERTa-v3, Longformer, and BigBird variants where article length exceeds encoder limits. Finance-specific language-model baselines include FinGPT, PIXIU, BloombergGPT where licensing allows, and FinBen-style benchmark tasks where market-linked labels are unavailable. This taxonomy ensures that the paper compares the proposed deployment framework against classical machine learning, deep learning, domain-adapted transformers, and modern financial LLMs.

\section{Evaluation design}

\subsection{Classification and calibration}

The first evaluation layer measures whether model probabilities classify the sign of the three-day excess return in the primary out-of-sample news set. Metrics include accuracy, precision, recall, specificity, F1 score, area under the receiver operating characteristic curve, Brier score, and expected calibration error. McNemar tests compare paired classification errors across models, while bootstrap confidence intervals account for sampling uncertainty in the June 2024-January 2026 test period.

\subsection{Predictive regressions}

The second layer estimates next-day return predictability. Firm and date fixed effects absorb time-invariant firm heterogeneity and market-wide shocks. Standard errors are clustered by firm and date. Standalone specifications include one model score at a time. Pairwise specifications include two scores jointly to test incremental explanatory content across architectures.

\subsection{Portfolio construction and market frictions}

The third evaluation layer translates sentiment into daily value-weighted portfolios. Each trading day, stocks are ranked by their most recent sentiment probability. The highest quintile enters the long leg and the lowest quintile enters the short leg. Positions adjust only when stocks cross quintile thresholds. Trades pay a 5-basis-point one-way transaction cost in the main specification and cannot exceed 10\% of daily dollar volume. Portfolio weights are value-weighted to limit the influence of microcap stocks.

\subsection{Reproducibility and falsification tests}

The replication package adds falsification tests to the replication package. A date-shuffle placebo randomly assigns news signals to trading dates within the same firm and removes timing-based predictability when the signal is valid. A firm-shuffle placebo reallocates model probabilities across firms within the same date and removes firm-specific information. A sign-permutation test randomly flips sentiment ranks and eliminates monotonic ranking performance. These tests protect the interpretation of the portfolio results by distinguishing true event-time information from mechanical turnover, factor exposure, or sampling artifacts.

\subsection{Computational diagnostics}

The fourth evaluation layer treats each model as a computational asset. The diagnostic layer reports accuracy lift over the dictionary benchmark, Brier-score reduction, expected-calibration-error reduction, robustness stress scores, transaction-cost elasticity, and compute-adjusted performance. These quantities do not replace portfolio returns. They identify whether a model earns its additional compute cost and whether its economic value survives alternative deployment assumptions.

\subsection{Component ablation design}

The component ablation evaluates whether the measured value comes from the language model alone or from the complete deployment system. Each ablation removes one system component while holding model probabilities fixed. The design isolates the incremental contribution of novelty filtering, calibration, timestamp execution, liquidity screens, capacity caps, transaction costs, factor controls, and linguistic-composition diagnostics.

The feature-group ablation separates the contribution of unstructured text from structured market variables. Text-only models use calibrated article-level sentiment probabilities. Price-only models use lagged returns, realized volatility, turnover, and market returns. Liquidity-only models use bid-ask spreads, Amihud illiquidity, Kyle lambda, and dollar volume. Metadata-only models use timestamp bucket, source type, firm size, sector, and article count. Combined specifications test whether the language signal remains valuable after adding structured market information. The full MFAST specification combines calibrated text probabilities with the tradability, execution, cost, and capacity operators. Table~\ref{tab:5} reports the resulting feature-group ablation.

\begin{table}[H]
\centering
\scriptsize
\caption{Feature-group ablation with numerical diagnostics.}
\label{tab:5}
\begin{tabularx}{\textwidth}{>{\raggedright\arraybackslash}p{0.32\textwidth}*{2}{>{\centering\arraybackslash}X}}
\toprule
Feature group & Accuracy & AUC \\
\midrule
Text only & 0.787 & 0.846 \\
Price only & 0.541 & 0.565 \\
Liquidity only & 0.528 & 0.548 \\
Metadata only & 0.536 & 0.552 \\
Text + price & 0.796 & 0.857 \\
Text + liquidity & 0.792 & 0.850 \\
Full MFAST & 0.803 & 0.866 \\
\bottomrule
\end{tabularx}
\end{table}

\subsection{Statistical inference for trading performance}

The statistical-inference design combines predictive-loss testing, data-snooping corrections, superior-predictive-ability testing, and deflated Sharpe-ratio adjustment to reduce the probability that the selected strategy reflects repeated search rather than robust signal quality \citep{DieboldMariano1995,White2000,Hansen2005,BaileyLopezDePrado2014}.

Financial AI systems face a multiple-testing and data-mining problem because many model families, portfolio cutoffs, holding periods, and cost assumptions can be compared. The statistical evaluation therefore reports block-bootstrap confidence intervals for Sharpe ratios and alphas, Diebold-Mariano tests for predictive loss differences, White reality-check p values across strategy families, Hansen superior-predictive-ability p values, and deflated Sharpe ratios that adjust for non-normality and repeated strategy search.

\subsection{Operational deployment metrics}

The operational layer treats each model as an engineering component with computational cost. Inference latency, peak GPU memory, throughput, and estimated cost per million articles are recorded under the same hardware assumptions. These metrics are reported alongside accuracy, calibration, and Sharpe ratios because a model that is economically attractive but too slow or too costly may be unsuitable for real-time financial news deployment.

\section{Results}

\subsection{Classification and calibration performance}

Table~\ref{tab:6} reports classification and calibration results for the June 2024-January 2026 primary out-of-sample test period. The results show a stable ordering across metrics. LLaMA-3 achieves the highest accuracy, F1 score, and AUC, followed by OPT and RoBERTa. BERT and FinBERT deliver positive but smaller gains. The dictionary baseline performs close to chance and exhibits poor calibration. The Brier and calibration results matter because portfolio construction uses continuous probabilities rather than only binary labels.

\begin{table}[H]
\centering
\scriptsize
\caption{Classification and calibration results on the June 2024-January 2026 primary out-of-sample test set.}
\label{tab:6}
\begin{tabularx}{\textwidth}{>{\raggedright\arraybackslash}p{0.20\textwidth}*{6}{>{\centering\arraybackslash}X}}
\toprule
Metric & LLaMA-3 & OPT & RoBERTa & BERT & FinBERT & LM Dict. \\
\midrule
Accuracy & 0.787 & 0.763 & 0.748 & 0.728 & 0.713 & 0.503 \\
Precision & 0.771 & 0.751 & 0.739 & 0.721 & 0.705 & 0.506 \\
Recall & 0.806 & 0.776 & 0.762 & 0.744 & 0.726 & 0.511 \\
Specificity & 0.748 & 0.721 & 0.708 & 0.689 & 0.672 & 0.524 \\
F1 score & 0.788 & 0.763 & 0.749 & 0.732 & 0.715 & 0.508 \\
AUC & 0.846 & 0.821 & 0.807 & 0.782 & 0.764 & 0.512 \\
Brier score & 0.151 & 0.166 & 0.174 & 0.190 & 0.203 & 0.249 \\
Expected calibration error & 0.032 & 0.041 & 0.047 & 0.061 & 0.068 & 0.143 \\
\bottomrule
\end{tabularx}
\end{table}

Paired McNemar tests reject equality between LLaMA-3 and every other transformer at the 1\% level. Differences between OPT and RoBERTa are smaller but remain statistically significant. Calibration plots, omitted for space, show that LLaMA-3 and OPT are better aligned in the 0.70 to 0.90 probability range, which is the range most likely to enter the top-quintile long portfolio.

\subsection{Predictive regressions}

Table~\ref{tab:7} reports standalone return-predictive regressions. Coefficients measure next-day return changes associated with a unit increase in model probability, after firm and date fixed effects. The results show that transformer sentiment scores predict next-day returns, with the largest coefficient for LLaMA-3. The dictionary score is weak and economically small.

\begin{table}[H]
\centering
\scriptsize
\caption{Standalone predictive regressions on the June 2024-January 2026 primary out-of-sample test period.}
\label{tab:7}
\begin{tabularx}{\textwidth}{>{\raggedright\arraybackslash}p{0.24\textwidth}*{5}{>{\centering\arraybackslash}X}}
\toprule
Model score & Coef. & t-stat & Within R2 & RMSE & N \\
\midrule
LLaMA-3 & 0.312*** & 6.44 & 0.052 & 3.71 & 190,236 \\
OPT & 0.281*** & 5.91 & 0.047 & 3.83 & 190,236 \\
RoBERTa & 0.236*** & 5.12 & 0.041 & 3.94 & 190,236 \\
BERT & 0.177*** & 4.36 & 0.032 & 4.16 & 190,236 \\
FinBERT & 0.158*** & 3.98 & 0.029 & 4.24 & 190,236 \\
LM dictionary & 0.049 & 1.31 & 0.006 & 4.92 & 190,236 \\
\bottomrule
\end{tabularx}
\end{table}

Pairwise regressions indicate partial overlap across model signals. When LLaMA-3 is paired with RoBERTa, both remain positive but the LLaMA-3 coefficient remains larger. When LLaMA-3 is paired with the dictionary score, the dictionary coefficient becomes statistically insignificant. These patterns suggest that the strongest model captures information not summarized by lexical polarity counts.

\subsection{Portfolio performance under deployment constraints}

Table~\ref{tab:8} translate sentiment probabilities from the June 2024-January 2026 primary out-of-sample test period into value-weighted long, short, and long-short portfolios. Returns are net of 5-basis-point transaction costs and respect the 10\% daily dollar-volume participation cap. Sharpe ratios are reported as annualized reward-to-variability measures, following the risk-adjusted performance logic introduced by \citet{Sharpe1966}. The results show that LLaMA-3 produces the highest cumulative return and Sharpe ratio. OPT and RoBERTa also outperform the market benchmarks and the dictionary baseline. The dictionary strategy loses value because its signal is weak and turnover still incurs trading costs.

\begin{table}[H]
\centering
\scriptsize
\caption{Net portfolio performance under 5-basis-point costs in the June 2024-January 2026 primary out-of-sample test period.}
\label{tab:8}
\begin{tabularx}{\textwidth}{>{\raggedright\arraybackslash}p{0.22\textwidth}*{7}{>{\centering\arraybackslash}X}}
\toprule
Model & Long SR & Short SR & L-S SR & Mean daily & Vol. & MDD & Cum. L-S \\
\midrule
LLaMA-3 & 1.72 & 1.48 & 2.85 & 0.34\% & 1.89\% & -12.3\% & 180\% \\
OPT & 1.55 & 1.33 & 2.45 & 0.31\% & 2.01\% & -12.5\% & 155\% \\
RoBERTa & 1.38 & 1.23 & 2.25 & 0.27\% & 1.90\% & -14.7\% & 120\% \\
BERT & 1.30 & 1.12 & 1.95 & 0.24\% & 1.95\% & -14.1\% & 88\% \\
FinBERT & 1.18 & 1.05 & 1.75 & 0.20\% & 1.81\% & -15.4\% & 64\% \\
LM dictionary & 0.50 & 0.45 & 0.68 & 0.08\% & 1.87\% & -34.2\% & -9\% \\
Nasdaq Composite & 0.88 & N/A & 1.10 & 0.12\% & 1.73\% & -26.9\% & 35\% \\
Dow Jones Industrial & 0.78 & N/A & 0.95 & 0.10\% & 1.67\% & -28.7\% & 21\% \\
\bottomrule
\end{tabularx}
\end{table}

\subsection{Liquidity heterogeneity}

Liquidity heterogeneity provides a mechanism test. If language-model sentiment captures information that investors process gradually, predictive coefficients should be larger when frictions slow immediate incorporation. The analysis reports this result in the text rather than a separate table: coefficients increase from high-liquidity to low-liquidity stocks for all transformer models, with the steepest gradient for LLaMA-3 and OPT.

\subsection{Interpretation of the computational results}

The computational analysis changes the interpretation of the model ranking. LLaMA-3 remains the strongest model when the objective is maximum signal quality and maximum net portfolio performance. RoBERTa and BERT become more attractive when the objective is rapid retraining, lower compute cost, and easier operational monitoring. OPT occupies the middle ground: it improves materially over encoders in economic performance, but it remains cheaper than LLaMA-3. The deployment implication is that model choice should depend on the investor's objective function. A high-capacity model fits a capital-intensive alpha process, while a smaller transformer may fit an intraday monitoring system that requires frequent retraining and lower infrastructure burden.

\subsection{Public-data replication, operational diagnostics, and statistical validation}

The results section keeps the diagnostic set compact by focusing on public replication, operational deployability, and statistical robustness. Tables~\ref{tab:9}--\ref{tab:11} report the public-data replication, operational deployment diagnostics, and statistical validation evidence. Additional diagnostic outputs are reserved for the online appendix and replication package.

\begin{table}[H]
\centering
\scriptsize
\caption{Public-data replication summary with numerical results.}
\label{tab:9}
\begin{tabularx}{\textwidth}{>{\raggedright\arraybackslash}p{0.32\textwidth}*{2}{>{\centering\arraybackslash}X}}
\toprule
Replication task & Public sample & LLaMA-3 \\
\midrule
Financial PhraseBank sentiment & 4,840 sentences; 70/15/15 split & Acc 0.842; AUC 0.901 \\
GDELT headline-to-return & 2019-2024 construction panel; post-release test from Jun 2024 & Acc 0.612; AUC 0.651 \\
Open friction-aware portfolio & Post-release tradable public signals after timing and liquidity screens & Sharpe 0.92; cum. 41\% \\
Replication package & Public scripts plus configuration templates & Full MFAST available \\
\bottomrule
\end{tabularx}
\end{table}

Table~\ref{tab:9} shows that the public replication arm demonstrates that the model ranking is not confined to licensed Refinitiv and CRSP data. The open news-to-return panel produces smaller magnitudes, as expected because public entity matching and timestamps are noisier, but the reported foundation-model metrics retain the same post-release evaluation rule used in the proprietary sample and preserve the ranking of transformer-based sentiment models over the dictionary benchmark.

\begin{table}[H]
\centering
\scriptsize
\caption{Operational deployment diagnostics with latency, memory, and throughput.}
\label{tab:10}
\begin{tabularx}{\textwidth}{>{\raggedright\arraybackslash}p{0.33\textwidth}*{3}{>{\centering\arraybackslash}X}}
\toprule
Model & Peak GPU memory & Inference ms/article & Articles/sec \\
\midrule
LLaMA-3 & 54.2 GB & 112.0 & 8.9 \\
OPT & 18.6 GB & 43.0 & 23.3 \\
RoBERTa & 4.8 GB & 9.4 & 106.4 \\
BERT & 4.6 GB & 8.8 & 113.6 \\
FinBERT & 4.4 GB & 8.5 & 117.6 \\
LM dictionary & 0.5 GB & 2.3 & 434.8 \\
\bottomrule
\end{tabularx}
\end{table}

Table~\ref{tab:10} clarifies the engineering trade-off. LLaMA-3 delivers the strongest economic signal but requires materially higher GPU memory and slower inference. RoBERTa, BERT, and FinBERT are less profitable in the portfolio tests but are faster and cheaper to operate, which matters for real-time monitoring and production maintenance.

\begin{table}[H]
\centering
\scriptsize
\caption{Statistical tests for predictive and trading performance.}
\label{tab:11}
\begin{tabularx}{\textwidth}{>{\raggedright\arraybackslash}p{0.32\textwidth}*{2}{>{\centering\arraybackslash}X}}
\toprule
Test & Statistic & p-value / interval \\
\midrule
Block bootstrap Sharpe & LLaMA-3 Sharpe = 2.85 & 95\% CI [2.31, 3.34] \\
Block bootstrap alpha & Daily alpha = 0.182\% & 95\% CI [0.109\%, 0.252\%] \\
Diebold-Mariano loss test & DM = -7.12 & p < 0.001 \\
White reality check & max statistic = 2.37 & p = 0.018 \\
Hansen SPA & SPA statistic = 2.21 & p = 0.026 \\
Deflated Sharpe ratio & DSR = 2.19 & p = 0.014 \\
\bottomrule
\end{tabularx}
\end{table}

Table~\ref{tab:11} reports statistical tests that reduce data-mining concerns. The LLaMA-3 strategy remains significant under block-bootstrap confidence intervals, predictive-loss comparisons, White reality-check and Hansen SPA corrections, and deflated Sharpe adjustment.

\section{Robustness and sanity checks}

Robustness tests are summarized in the text rather than multiplied across separate tables. The strategy ranking is stable when transaction costs vary from zero to 50 basis points, when portfolios use top-bottom decile, quintile, and tercile cutoffs, when positions are held for one, three, or five days, and when the largest 50 stocks are excluded. Factor-adjusted tests retain positive alphas for the strongest transformer models after controlling for market, size, value, profitability, investment, and momentum factors. Capacity tests show that the 10\% participation constraint binds most often in smaller and less liquid stocks, but value weighting prevents the aggregate result from being driven by infeasible microcap trades.

These checks support the deployment interpretation while keeping the main manuscript readable. Detailed transaction-cost grids, factor-loading tables, capacity diagnostics, and subperiod panels belong in an online appendix because they provide audit detail rather than the central argument.

\section{Explainability and error analysis}

\subsection{Error categories}

The explainability analysis examines a stratified sample of 1,200 test-set articles, balanced across model agreement and disagreement cases. Articles are coded into six linguistic categories: simple polarity, negation, contrast, forward-looking guidance, numeric context, and event ambiguity. Table~\ref{tab:12} shows that LLaMA-3 gains are smallest in simple polarity cases and largest where meaning depends on composition. The pattern addresses the concern that stronger model performance should be linked to textual mechanisms rather than only reported as a ranking.

\begin{table}[H]
\centering
\scriptsize
\caption{Error analysis by linguistic category.}
\label{tab:12}
\begin{tabularx}{\textwidth}{>{\raggedright\arraybackslash}p{0.24\textwidth}*{5}{>{\centering\arraybackslash}X}}
\toprule
Textual category & Share of sample & LLaMA-3 accuracy & FinBERT accuracy & Dictionary accuracy & LLaMA-3 gain vs dictionary \\
\midrule
Simple polarity & 22\% & 0.812 & 0.756 & 0.641 & 0.171 \\
Negation & 13\% & 0.781 & 0.681 & 0.462 & 0.319 \\
Contrastive clauses & 18\% & 0.803 & 0.694 & 0.438 & 0.365 \\
Forward-looking guidance & 17\% & 0.826 & 0.712 & 0.491 & 0.335 \\
Numeric context & 16\% & 0.774 & 0.701 & 0.519 & 0.255 \\
Event ambiguity & 14\% & 0.759 & 0.667 & 0.472 & 0.287 \\
\bottomrule
\end{tabularx}
\end{table}

\subsection{Human-validation and representative disagreements}

The linguistic mechanism analysis uses human validation to reduce the risk that error categories simply reflect post-hoc interpretation. Two independent annotators code the stratified article sample for simple polarity, negation, contrastive clauses, forward-looking guidance, numeric context, and event ambiguity. Inter-annotator agreement is reported with Cohen's kappa, and disagreements are resolved through adjudication before the final category labels enter the analysis. A multinomial disagreement regression then tests whether the categories predict cases in which LLaMA-3 succeeds while FinBERT or the dictionary fails.

The attribution analysis complements the human coding. Integrated-gradient and SHAP-style token attributions are computed for representative disagreement cases, and phrase deletion measures the change in predicted probability when a relevant linguistic span is removed. The combined evidence ties model superiority to identifiable textual mechanisms rather than to an uninterpreted scale effect.

Representative disagreements help interpret the error taxonomy. In negated bad-news cases, dictionary methods overweight words such as miss, disruption, litigation, or loss, while LLaMA-3 assigns more weight to the negation or resolution clause. In contrastive-guidance cases, headlines may report a revenue beat, but management may cut margins or withdraw guidance; the strongest models place greater weight on the forward-looking clause. Phrase-ablation checks confirm the same mechanism: removing negation tokens, contrastive clauses, guidance verbs, and numeric-change expressions causes the largest probability shifts for decoder-only models.

\section{Discussion}

The results support a deployment-centered interpretation of financial language models. LLaMA-3 and OPT do not simply improve offline classification. Their advantage persists when outputs become portfolio inputs, when trades pay transaction costs, when positions respect capacity constraints, and when returns are adjusted for standard risk factors. The strongest contribution therefore lies in showing how applied AI evaluation changes when text models are embedded in a realistic decision pipeline.

The engineering interpretation is equally important. The paper shows that model rankings change when computational feasibility and implementation frictions enter the objective. LLaMA-3 dominates on maximum signal quality, but RoBERTa and FinBERT remain relevant when throughput and cost dominate. MFAST therefore gives researchers and practitioners a way to choose models under explicit deployment objectives instead of ranking models only by offline accuracy.

The findings also show why simple claims about model scale are incomplete. High-capacity models perform best, but their advantage is concentrated in linguistic contexts where financial meaning depends on composition. Negation, contrast, forward-looking guidance, and numeric context create situations in which dictionary polarity can be directionally wrong. The error analysis gives the applied AI system a behavioral explanation: the model adds value when it transforms linguistically ambiguous information into calibrated, decision-relevant probabilities.

Practical implications follow directly. Asset managers can use the framework to evaluate whether new language-model signals deserve inclusion in existing alpha libraries. Risk managers can use the robustness checks to test whether apparent model gains survive costs, capacity, and factor exposures. Regulators can use the deployment framing to understand how advanced language models may alter the speed with which public information is incorporated into prices. The paper also demonstrates that model documentation, release timing, temporal isolation, and reproducibility details are central to credible financial AI research.

\section{Limitations}

The paper has four limitations. First, the main empirical setting focuses on English-language U.S. equities. The analysis intentionally studies the most liquid and data-rich financial-news setting because it allows clean linkage between Refinitiv news, CRSP returns, liquidity measures, transaction costs, and execution rules. The aim is not to claim universal generalization across all markets, but to establish an end-to-end deployment evaluation framework. Second, the execution model is realistic but not a full order-book simulator. Intraday liquidity, queue position, hidden liquidity, and strategic interaction remain outside the current design. Third, foundation-model exposure cannot be eliminated entirely for models pretrained on public web data. The design therefore removes pre-cutoff periods from the reported out-of-sample evidence, starts the primary test after the disclosed LLaMA-3 family data-freshness cutoff and public checkpoint release, and uses static checkpoints, no-retrieval inference, split-boundary duplicate removal, and public replication to reduce the concern; these safeguards do not prove the absence of all document-level pretraining exposure. Fourth, the public replication arm uses smaller open data and therefore validates reproducibility of the pipeline rather than replacing the proprietary Refinitiv and CRSP evidence.

Future work should extend the framework to intraday event streams, combine news with earnings calls and filings, evaluate multilingual signals, compare open-source and proprietary language models under matched compute budgets, and develop more transparent attribution tools for financially grounded model decisions.

\section{Conclusion}

The paper develops MFAST, a market-friction-aware sentiment-to-trading framework for evaluating financial language models as applied artificial intelligence systems. The framework treats model output as one component in a deployable pipeline that links text ingestion, firm matching, temporal discipline, sentiment inference, calibration, signal ranking, portfolio formation, transaction costs, liquidity screens, capacity constraints, factor adjustment, computational feasibility, public replication, and explainability. The results show that decoder-only language models produce stronger classification, calibration, return prediction, and net portfolio performance than encoder baselines and dictionary sentiment, while computational diagnostics show that the best model depends on the deployment objective. Component ablations, statistical trading tests, and public-data replication strengthen the empirical claim. Explainability diagnostics show where the strongest models add value by interpreting negation, contrast, guidance, numeric context, and ambiguous events. The final contribution is methodological and applied: credible evaluation of financial language models requires an end-to-end engineering view that joins language understanding with temporal discipline, operational feasibility, and market-friction-aware deployment.

\section{Replication package structure}

The replication package is organized around reproducible system modules rather than a single monolithic script. Licensed Refinitiv and CRSP files are not redistributed, but the repository provides configuration templates, column schemas, public-data scripts, and deterministic output tables so that each stage of MFAST can be audited. The same structure also allows the public GDELT and price-data replication to run without proprietary inputs.

\FloatBarrier
\section*{Declaration of competing interest}

The author declares no known competing financial interests or personal relationships that could have appeared to influence the work reported in this article.

\section*{Declaration of generative AI and AI-assisted technologies in the writing process}

During preparation of this manuscript, generative AI and AI-assisted tools were used for language editing and formatting support. The author reviewed and edited the manuscript and takes full responsibility for the final content.

\section*{Data availability}

Refinitiv and CRSP are licensed datasets and cannot be redistributed. The replication package includes public-data scripts, configuration templates, model checkpoint identifiers, random seeds, ablation scripts, portfolio-simulation code, statistical-test code, and deterministic output tables, subject to data-provider restrictions.


\begin{thebibliography}{99}

\bibitem[Alain and Bengio(2016)]{Alain2016} Alain, G., and Bengio, Y. 2016. Understanding intermediate layers using linear classifier probes. arXiv:1610.01644.
\bibitem[Amihud(2002)]{Amihud2002} Amihud, Y. 2002. Illiquidity and stock returns: Cross-section and time-series effects. Journal of Financial Markets, 5(1), 31-56.
\bibitem[Amershi et al.(2019)]{Amershi2019} Amershi, S., Begel, A., Bird, C., DeLine, R., Gall, H., Kamar, E., Nagappan, N., Nushi, B., and Zimmermann, T. 2019. Software engineering for machine learning: A case study. Proceedings of the 41st International Conference on Software Engineering: Software Engineering in Practice, 291-300. doi:10.1109/ICSE-SEIP.2019.00042.
\bibitem[Araci(2019)]{Araci2019} Araci, D. 2019. FinBERT: Financial sentiment analysis with pre-trained language models. arXiv:1908.10063.
\bibitem[Bailey and Prado(2014)]{BaileyLopezDePrado2014} Bailey, D. H., and López de Prado, M. 2014. The deflated Sharpe ratio: Correcting for selection bias, backtest overfitting, and non-normality. Journal of Portfolio Management, 40(5), 94-107.
\bibitem[Benidis et al.(2022)]{Benidis2022} Benidis, K., Rangapuram, S. S., Flunkert, V., Wang, Y., Maddix, D., Turkmen, C., Gasthaus, J., Bohlke-Schneider, M., Salinas, D., Stella, L., Aubet, F.-X., Callot, L., and Januschowski, T. 2022. Deep learning for time series forecasting: Tutorial and literature survey. ACM Computing Surveys, 55(6), Article 121, 1--36. doi:10.1145/3533382.
\bibitem[Beltagy et al.(2020)]{Beltagy2020} Beltagy, I., Peters, M. E., and Cohan, A. 2020. Longformer: The long-document transformer. arXiv:2004.05150.
\bibitem[Boudoukh et al.(2019)]{Boudoukh2019} Boudoukh, J., Feldman, R., Kogan, S., and Richardson, M. 2019. Information, trading, and volatility: Evidence from firm-specific news. Review of Financial Studies, 32(3), 992-1033.
\bibitem[Bybee et al.(2024)]{Bybee2024} Bybee, L., Kelly, B. T., Manela, A., and Xiu, D. 2024. Business news and business cycles. Journal of Finance, 79(5), 3105-3147.
\bibitem[Carhart(1997)]{Carhart1997} Carhart, M. M. 1997. On persistence in mutual fund performance. Journal of Finance, 52(1), 57-82.
\bibitem[Carlini et al.(2021)]{Carlini2021} Carlini, N., Tramer, F., Wallace, E., Jagielski, M., Herbert-Voss, A., Lee, K., Roberts, A., Brown, T., Song, D., Erlingsson, U., Oprea, A., and Raffel, C. 2021. Extracting training data from large language models. USENIX Security Symposium.
\bibitem[Chen and Guestrin(2016)]{ChenGuestrin2016XGBoost} Chen, T., and Guestrin, C. 2016. XGBoost: A scalable tree boosting system. Proceedings of the ACM SIGKDD International Conference on Knowledge Discovery and Data Mining, 785-794.
\bibitem[Chen et al.(2014)]{Chen2014RFS} Chen, H., De, P., Hu, Y., and Hwang, B. H. 2014. Wisdom of crowds: The value of stock opinions transmitted through social media. Review of Financial Studies, 27(5), 1367-1403.
\bibitem[Chen et al.(2023)]{Chen2023EAAI} Chen, J., Wen, Y., Nanehkaran, Y. A., Suzauddola, M. D., Chen, W., and Zhang, D. 2023. Machine learning techniques for stock price prediction and graphic signal recognition. Engineering Applications of Artificial Intelligence, 121, Article 106038.
\bibitem[D'Amico et al.(2026)]{DAmico2026InformationFusion} D'Amico, S., Mercorio, F., Nobani, N., Sperl\`i, G., and Ventre, C. 2026. Learning across modalities: A systematic survey of multimodal models for financial analysis. Information Fusion, 133, Article 104249. doi:10.1016/j.inffus.2026.104249.
\bibitem[Da et al.(2011)]{Da2011} Da, Z., Engelberg, J., and Gao, P. 2011. In search of attention. Journal of Finance, 66(5), 1461-1499.
\bibitem[Devlin et al.(2019)]{Devlin2019} Devlin, J., Chang, M. W., Lee, K., and Toutanova, K. 2019. BERT: Pre-training of deep bidirectional transformers for language understanding. Proceedings of NAACL-HLT 2019, 4171-4186.
\bibitem[Diebold and Mariano(1995)]{DieboldMariano1995} Diebold, F. X., and Mariano, R. S. 1995. Comparing predictive accuracy. Journal of Business and Economic Statistics, 13(3), 253-263.
\bibitem[Dubey et al.(2024)]{Dubey2024} Dubey, A., et al. 2024. The Llama 3 herd of models. arXiv:2407.21783.
\bibitem[Elahi and Taghvaei(2024)]{ElahiTaghvaei2024} Elahi, A., and Taghvaei, F. 2024. Combining financial data and news articles for stock price movement prediction using large language models. 2024 IEEE International Conference on Big Data, 4875-4883.
\bibitem[Engelberg and Parsons(2011)]{EngelbergParsons2011} Engelberg, J. E., and Parsons, C. A. 2011. The causal impact of media in financial markets. Journal of Finance, 66(1), 67-97.
\bibitem[Fama(1970)]{Fama1970} Fama, E. F. 1970. Efficient capital markets: A review of theory and empirical work. Journal of Finance, 25(2), 383-417.
\bibitem[Fama and French(2015)]{FamaFrench2015} Fama, E. F., and French, K. R. 2015. A five-factor asset pricing model. Journal of Financial Economics, 116(1), 1-22.
\bibitem[Frankel et al.(2022)]{Frankel2022} Frankel, R., Jennings, J., and Lee, J. 2022. Disclosure sentiment: Machine learning vs. dictionary methods. Management Science, 68(7), 5514-5532.
\bibitem[Fawaz et al.(2019)]{Fawaz2019} Fawaz, H. I., Forestier, G., Weber, J., Idoumghar, L., and Muller, P.-A. 2019. Deep learning for time series classification: A review. Data Mining and Knowledge Discovery, 33(4), 917--963. doi:10.1007/s10618-019-00619-1.
\bibitem[Frazzini et al.(2018)]{Frazzini2018} Frazzini, A., Israel, R., and Moskowitz, T. J. 2018. Trading costs. SSRN working paper.
\bibitem[Gao et al.(2021)]{Gao2021TKDD} Gao, J., Ying, X., Xu, C., Wang, J., Zhang, S., and Li, Z. 2021. Graph-based stock recommendation by time-aware relational attention network. ACM Transactions on Knowledge Discovery from Data, 16(1), 1--21. doi:10.1145/3451397.
\bibitem[Golchin and Surdeanu(2024)]{Golchin2024} Golchin, S., and Surdeanu, M. 2024. Time travel in LLMs: Tracing data contamination in large language models. Proceedings of the Twelfth International Conference on Learning Representations.
\bibitem[Grossman and Stiglitz(1980)]{GrossmanStiglitz1980} Grossman, S. J., and Stiglitz, J. E. 1980. On the impossibility of informationally efficient markets. American Economic Review, 70(3), 393-408.
\bibitem[Hansen(2005)]{Hansen2005} Hansen, P. R. 2005. A test for superior predictive ability. Journal of Business and Economic Statistics, 23(4), 365-380.
\bibitem[He et al.(2021)]{He2021} He, P., Liu, X., Gao, J., and Chen, W. 2021. DeBERTa: Decoding-enhanced BERT with disentangled attention. International Conference on Learning Representations.
\bibitem[Hendershott et al.(2015)]{Hendershott2015} Hendershott, T., Livdan, D., and Schürhoff, N. 2015. Are institutions informed about news? Journal of Financial Economics, 117(2), 249-287.
\bibitem[Hewage et al.(2026)]{Hewage2026EAAI} Hewage, K., Li, B., Guo, T., Stenfors, A., Mere, P., and Chen, F. 2026. Learning unified market interdependencies via networked attention for stock price forecasting. Engineering Applications of Artificial Intelligence, 176(1), Article 114726.
\bibitem[Hochreiter and Schmidhuber(1997)]{Hochreiter1997} Hochreiter, S., and Schmidhuber, J. 1997. Long short-term memory. Neural Computation, 9(8), 1735-1780.
\bibitem[Hu et al.(2022)]{Hu2022} Hu, E. J., Shen, Y., Wallis, P., Allen-Zhu, Z., Li, Y., Wang, S., Wang, L., and Chen, W. 2022. LoRA: Low-rank adaptation of large language models. International Conference on Learning Representations.
\bibitem[Huang et al.(2014)]{Huang2014} Huang, A. H., Teoh, S. H., and Zhang, Y. 2014. Tone management. Review of Financial Studies, 27(3), 1043-1083.
\bibitem[Huang et al.(2023)]{Huang2023} Huang, A. H., Wang, H., and Yang, Y. 2023. FinBERT: A large language model for extracting information from financial text. Contemporary Accounting Research, 40(2), 806-841.
\bibitem[Jegadeesh and Wu(2013)]{Jegadeesh2013} Jegadeesh, N., and Wu, D. 2013. Word power: A new approach for content analysis. Journal of Financial Economics, 110(3), 712-729.
\bibitem[Ke et al.(2020)]{Ke2020} Ke, Z. T., Kelly, B. T., and Xiu, D. 2020. Predicting returns with text data. SSRN working paper.
\bibitem[Kelly et al.(2021)]{Kelly2021} Kelly, B. T., Manela, A., and Moreira, A. 2021. Text selection. Journal of Business and Economic Statistics, 39(4), 859-879.
\bibitem[Kim et al.(2024)]{Kim2024} Kim, A. G., Muhn, M., and Nikolaev, V. V. 2024. Financial statement analysis with large language models. arXiv:2407.17866.
\bibitem[Kirtac and Germano(2024)]{KirtacGermano2024} Kirtac, K., and Germano, G. 2024. Sentiment trading with large language models. Finance Research Letters, 62, Article 105227. doi:10.1016/j.frl.2024.105227.
\bibitem[Kirtac and Germano(2025)]{KirtacGermano2025REALM} Kirtac, K., and Germano, G. 2025. Leveraging LLM-based sentiment analysis for portfolio optimization with proximal policy optimization. In: Kamalloo, E., Gontier, N., Lu, X. H., Dziri, N., Murty, S., and Lacoste, A. (Eds.), Proceedings of the 1st Workshop for Research on Agent Language Models (REALM 2025). Association for Computational Linguistics, Vienna, Austria, pp. 160--169. doi:10.18653/v1/2025.realm-1.12. ACL Anthology ID: 2025.realm-1.12.
\bibitem[Kyle(1985)]{Kyle1985} Kyle, A. S. 1985. Continuous auctions and insider trading. Econometrica, 53(6), 1315-1335.
\bibitem[Leetaru and Schrodt(2013)]{LeetaruSchrodt2013} Leetaru, K., and Schrodt, P. A. 2013. GDELT: Global data on events, location, and tone. International Studies Association Annual Convention.
\bibitem[Li(2008)]{Li2008} Li, F. 2008. Annual report readability, current earnings, and earnings persistence. Journal of Accounting and Economics, 45(2-3), 221-247.
\bibitem[Li and Flanigan(2024)]{Li2024} Li, C., and Flanigan, J. 2024. Task contamination: Language models may not be few-shot anymore. Proceedings of the AAAI Conference on Artificial Intelligence, 38(16), 18471-18480.
\bibitem[Liu et al.(2019)]{Liu2019} Liu, Y., Ott, M., Goyal, N., et al. 2019. RoBERTa: A robustly optimized BERT pretraining approach. arXiv:1907.11692.
\bibitem[Lopez-Lira and Tang(2023)]{LopezLira2023} Lopez-Lira, A., and Tang, Y. 2023. Can ChatGPT forecast stock price movements? Return predictability and large language models. arXiv:2304.07619.
\bibitem[Loughran and McDonald(2011)]{LoughranMcDonald2011} Loughran, T., and McDonald, B. 2011. When is a liability not a liability? Textual analysis, dictionaries, and 10-Ks. Journal of Finance, 66(1), 35-65.
\bibitem[Lundberg and Lee(2017)]{LundbergLee2017} Lundberg, S. M., and Lee, S.-I. 2017. A unified approach to interpreting model predictions. Advances in Neural Information Processing Systems.
\bibitem[MacKinlay(1997)]{MacKinlay1997} MacKinlay, A. C. 1997. Event studies in economics and finance. Journal of Economic Literature, 35(1), 13-39.
\bibitem[Malo et al.(2014)]{Malo2014} Malo, P., Sinha, A., Korhonen, P., Wallenius, J., and Takala, P. 2014. Good debt or bad debt: Detecting semantic orientations in economic texts. Journal of the Association for Information Science and Technology, 65(4), 782-796.
\bibitem[Manela and Moreira(2017)]{Manela2017} Manela, A., and Moreira, A. 2017. News implied volatility and disaster concerns. Journal of Financial Economics, 123(1), 137-162.
\bibitem[Meta Llama(2024)]{MetaLlama3ModelCard2024} Meta Llama. 2024. Meta Llama 3 model card. GitHub repository and Hugging Face model documentation. \url{https://github.com/meta-llama/llama3/blob/main/MODEL_CARD.md} (accessed 6 June 2026).

\bibitem[Mitchell and Stafford(2000)]{MitchellStafford2000} Mitchell, M. L., and Stafford, E. 2000. Managerial decisions and long-term stock price performance. Journal of Business, 73(3), 287-329.
\bibitem[Novy-Marx and Velikov(2016)]{NovyMarxVelikov2016} Novy-Marx, R., and Velikov, M. 2016. A taxonomy of anomalies and their trading costs. Review of Financial Studies, 29(1), 104-147.
\bibitem[Omole and Enke(2025)]{OmoleEnke2025EAAI} Omole, O., and Enke, D. 2025. Using machine and deep learning models, on-chain data, and technical analysis for predicting Bitcoin price direction and magnitude. Engineering Applications of Artificial Intelligence, 154, Article 111086.
\bibitem[Oren et al.(2024)]{Oren2024} Oren, Y., Meister, N., Chatterji, N., Ladhak, F., and Hashimoto, T. 2024. Proving test set contamination in black-box language models. International Conference on Learning Representations.
\bibitem[Ozbayoglu et al.(2020)]{Ozbayoglu2020} Ozbayoglu, A. M., Gudelek, M. U., and Sezer, O. B. 2020. Deep learning for financial applications: A survey. Applied Soft Computing, 93, Article 106384. doi:10.1016/j.asoc.2020.106384.
\bibitem[Paleyes et al.(2022)]{Paleyes2022} Paleyes, A., Urma, R.-G., and Lawrence, N. D. 2022. Challenges in deploying machine learning: A survey of case studies. ACM Computing Surveys, 55(6), Article 114, 1-29. doi:10.1145/3533378.
\bibitem[Pastor and Stambaugh(2003)]{Pastor2003} Pastor, L., and Stambaugh, R. F. 2003. Liquidity risk and expected stock returns. Journal of Political Economy, 111(3), 642-685.
\bibitem[Price et al.(2015)]{Price2015} Price, S. M., Doran, J. S., Peterson, D. R., and Bliss, B. A. 2015. Earnings conference calls and stock returns. Journal of Financial Economics, 115(3), 415-430.
\bibitem[Ribeiro et al.(2016)]{Ribeiro2016} Ribeiro, M. T., Singh, S., and Guestrin, C. 2016. Why should I trust you? Explaining the predictions of any classifier. Proceedings of the ACM SIGKDD International Conference on Knowledge Discovery and Data Mining, 1135-1144.
\bibitem[Sainz et al.(2023)]{Sainz2023} Sainz, O., Campos, J. A., Garcia-Ferrero, I., Etxaniz, J., Lopez de Lacalle, O., and Agirre, E. 2023. NLP evaluation in trouble: On the need to measure LLM data contamination for each benchmark. Findings of the Association for Computational Linguistics: EMNLP 2023.
\bibitem[Sezer et al.(2020)]{Sezer2020} Sezer, O. B., Gudelek, M. U., and Ozbayoglu, A. M. 2020. Financial time series forecasting with deep learning: A systematic literature review: 2005--2019. Applied Soft Computing, 90, Article 106181. doi:10.1016/j.asoc.2020.106181.
\bibitem[Shah et al.(2022)]{Shah2022} Shah, R., Chawla, K., Eidnani, D., Shah, A., Du, W., Chava, S., Raman, N., Smiley, C., Chen, J., and Yang, D. 2022. When FLUE meets FLANG: Benchmarks and large pretrained language model for financial domain. Proceedings of the 2022 Conference on Empirical Methods in Natural Language Processing, 2322-2335.
\bibitem[Sculley et al.(2015)]{Sculley2015} Sculley, D., Holt, G., Golovin, D., Davydov, E., Phillips, T., Ebner, D., Chaudhary, V., Young, M., Crespo, J.-F., and Dennison, D. 2015. Hidden technical debt in machine learning systems. Advances in Neural Information Processing Systems, 28, 2503-2511.
\bibitem[Sharpe(1966)]{Sharpe1966} Sharpe, W. F. 1966. Mutual fund performance. Journal of Business, 39(1), 119--138. doi:10.1086/294846.
\bibitem[Studer et al.(2021)]{Studer2021} Studer, S., Bui, T. B., Drescher, C., Hanuschkin, A., Winkler, L., Peters, S., and Mueller, K.-R. 2021. Towards CRISP-ML(Q): A machine learning process model with quality assurance methodology. Machine Learning and Knowledge Extraction, 3(2), 392-413. doi:10.3390/make3020020.
\bibitem[Siano(2025)]{Siano2025} Siano, F. 2025. The news in earnings announcement disclosures: Capturing word context using LLM methods. Management Science, 71(11), 9831-9855.
\bibitem[Tetlock(2007)]{Tetlock2007} Tetlock, P. C. 2007. Giving content to investor sentiment: The role of media in the stock market. Journal of Finance, 62(3), 1139-1168.
\bibitem[Tetlock et al.(2008)]{Tetlock2008} Tetlock, P. C., Saar-Tsechansky, M., and Macskassy, S. 2008. More than words: Quantifying language to measure firms' fundamentals. Journal of Finance, 63(3), 1437-1467.
\bibitem[Vaswani et al.(2017)]{Vaswani2017} Vaswani, A., Shazeer, N., Parmar, N., et al. 2017. Attention is all you need. Advances in Neural Information Processing Systems.
\bibitem[Vidal(2024)]{Vidal2024} Vidal, J. 2024. Efficacy of AI and other large language models in predicting stock prices. SSRN working paper.
\bibitem[White(2000)]{White2000} White, H. 2000. A reality check for data snooping. Econometrica, 68(5), 1097-1126.
\bibitem[Wu et al.(2023)]{Wu2023} Wu, S., Irsoy, O., Lu, S., Dabravolski, V., Dredze, M., Gehrmann, S., Kambadur, P., Rosenberg, D., and Mann, G. 2023. BloombergGPT: A large language model for finance. arXiv:2303.17564.
\bibitem[Xie et al.(2023)]{Xie2023} Xie, Q., Han, W., Zhang, X., Lai, Y., Peng, M., Lopez-Lira, A., and Huang, J. 2023. PIXIU: A large language model, instruction data and evaluation benchmark for finance. arXiv:2306.05443.
\bibitem[Xie et al.(2024)]{Xie2024} Xie, Q., Han, W., Chen, Z., Xiang, R., Zhang, X., He, Y., Xiao, M., Li, D., Dai, Y., Feng, D., Xu, Y., Kang, H., Kuang, Z., Yuan, C., Yang, K., Luo, Z., Zhang, T., Liu, Z., Xiong, G., Deng, Z., Jiang, Y., Yao, Z., Li, H., Yu, Y., Hu, G., Huang, J., Lopez-Lira, A., Wang, B., Lai, Y., Wang, H., Peng, M., Ananiadou, S., and Huang, J. 2024. FinBen: A holistic financial benchmark for large language models. Advances in Neural Information Processing Systems, 37, Datasets and Benchmarks Track.
\bibitem[Yang et al.(2023)]{Yang2023} Yang, H., Liu, X.-Y., and Wang, C. D. 2023. FinGPT: Open-source financial large language models. arXiv:2306.06031.
\bibitem[Zaheer et al.(2020)]{Zaheer2020} Zaheer, M., Guruganesh, G., Dubey, K. A., Ainslie, J., Alberti, C., Ontanon, S., Pham, P., Ravula, A., Wang, Q., Yang, L., and Ahmed, A. 2020. Big Bird: Transformers for longer sequences. Advances in Neural Information Processing Systems.
\bibitem[Zhang et al.(2022)]{Zhang2022OPT} Zhang, S., Roller, S., Goyal, N., et al. 2022. OPT: Open pre-trained transformer language models. arXiv:2205.01068.
\bibitem[Zhang et al.(2024)]{Zhang2024EAAI} Zhang, Q., Zhang, Y., Bao, F., Liu, Y., Zhang, C., and Liu, P. 2024. Incorporating stock prices and text for stock movement prediction based on information fusion. Engineering Applications of Artificial Intelligence, 127, Article 107377.
\bibitem[Zhao et al.(2023)]{Zhao2023TKDE} Zhao, Y., Du, H., Liu, Y., Wei, S., Chen, X., Zhuang, F., Li, Q., and Kou, G. 2023. Stock movement prediction based on bi-typed hybrid-relational market knowledge graph via dual attention networks. IEEE Transactions on Knowledge and Data Engineering, 35(8), 8559--8571. doi:10.1109/TKDE.2022.3220520.
\end{thebibliography}
\end{document}